\documentclass[sigconf]{acmart}
\usepackage[utf8]{inputenc}
\usepackage{graphicx}
\usepackage{subfigure}

\author{Jean-Thomas Baillargeon}
\email{jean-thomas.baillargeon@ift.ulaval.ca}
\affiliation{%
  \institution{Université Laval}
  \city{Québec}
  \country{Canada}
}

\author{Hélène Cossette}
\email{helene.cossette@act.ulaval.ca}
\affiliation{%
  \institution{Université Laval}
  \city{Québec}
  \country{Canada}
}

\author{Luc Lamontagne}
\email{luc.lamontagne@ift.ulaval.ca}
\affiliation{%
  \institution{Université Laval}
  \city{Québec}
  \country{Canada}
}

\renewcommand{\shortauthors}{Baillargeon et al.}
\title{Preventing RNN from Using Sequence Length as a Feature}

\begin{document}

\begin{abstract}
    Recurrent neural networks are deep learning topologies that can be trained to classify long documents. However, in our recent work, we found a critical problem with these cells: they can use the length differences between texts of different classes as a prominent classification feature. This has the effect of producing models that are brittle and fragile to concept drift, can provide misleading performances and are trivially explainable regardless of text content. This paper illustrates the problem using synthetic and real-world data and provides a simple solution using weight decay regularization.
\end{abstract}



\setcopyright{acmcopyright}
\copyrightyear{2022}
\acmYear{2022}

\acmConference[NLPIR '22]{}{December 16--18,
  2022}{Bangkok, Thailand}
\acmISBN{978-1-4503-9763-6}

\maketitle

\section{Introduction}
Neural networks powered by recurrent cells, such as Long Short-Term Memory (LSTM), have been used to construct state-of-the-art document classification models. Their theoretical capability of ingesting variable-length inputs into a machine learning pipeline has made them attractive to the scientific community. Even though models based on the Transformer architecture presented in \cite{vaswani2017attention} tend to offer improved performance for many natural language processing tasks, Recurrent Neural Networks (RNN) cells still act as a solid baseline for many tasks and should be treated accordingly. 

In our most recent work, we used RNN to classify insurance claims composed of many notes (i.e., textual documents) into benign or catastrophic. As we tried to extract the risk factors explaining the catastrophic claims, we discovered that the recurrent cells use input length as a predominant feature to classify claims. Upon investigation, we found that catastrophic and benign claims were composed of significantly different size entries: catastrophic claims contained an average of 300 documents, while benign claims contained an average of 50 documents. Although the length difference was observed during our preliminary data exploration, no previous work in the literature seems to warn against this potential behavior.

This feature may be considered legitimate for classification and performance purposes since the generalization over test sets with similar length distributions is good. However, we believe that using input lengths to fit a model makes it susceptible to concept drift, since a change in document length invalidates its classifications. This behavior complicates knowledge extraction using an attention or explainability mechanism, since the models' decision can be trivially explained by the length of the inputs.
Moreover this is a real problem that can go undetected in many real-world applications.

In this paper, we investigate the behavior of recurrent cells learning that tend to use sequence lengths as a feature and we explore some potential solutions. We explain in Section \ref{sec:related_work} that this problem has not yet been properly addressed in the literature. We illustrate the problem in Section \ref{sec:experimentation} and \ref{sec:extent_of_the_problem} using synthetic and real-world datasets. Finally, experimental results on how to deal with this problem are presented and analyzed in Section \ref{sec:results}.

Our contributions are twofold. First, we present the statistical red flags observed during our data investigation, and second, we present a solution to this problem by using adequate regularization. 

\section{Related Work}\label{sec:related_work}
Text document classification is a fundamental task in natural language processing (NLP). As more complex corpora became available, this generated a need to provide more flexible algorithms to handle the newly discovered problems.
A good example is the classification of long documents which requires an algorithm to model the long-term dependency between text elements in a sequence. 

An initial approach is to use an RNN Cell. As the name suggests, this cell recursively ingests the sequence of tokens from a document to generate a compact representation that can be used for different NLP tasks. These models have proven to be complex to train, as the long sequences cause the vanishing gradient problem, as discussed in \cite{pascanu2013difficulty}. Later implementations, such as LSTM \cite{hochreiter1997long}, allow better handling of gradients for modeling very long documents.

RNN's ability to count and maintain memory of long sequences is reasonably well documented. LSTMs can explicitly count the length of a sequence \cite{gers2000recurrent}, provide the next element in a counting sequence \cite{rodriguez1999recurrent} or count items in a picture \cite{fang2018can}. However, the literature has not addressed the problem that RNN can converge to a local minimum based on the sequence length difference in the classes.

Finally, most NLP state-of-the-art models now use the Transformer architecture \cite{vaswani2017attention}. To classify long documents, models that belong to the Longformer family (see \cite{beltagy2020longformer}) are used. As demonstrated by \cite{varivs2021sequence} for a sequence-to-sequence task, these models face an input length problem similar to the one we are studying. However, the problem remains to be investigated for classification tasks.

\section{Experimentation with Synthetic Data}\label{sec:experimentation}
We start by exploring the problem using synthetic data. To conduct this experiment, we generate datasets containing two classes with different sequence length distributions. By varying the gap between the distributions, we can evaluate its impact on the decisions of an RNN classification model. In this section, we present how we generate the input and how we evaluate the effect of the problem.

\subsection{Generation of the Synthetic Sequences} \label{subsec:simulated_sequence}
The creation of random datasets allows us to create uninformative inputs while controlling the sequence length meta-feature. A binary classification model trained on an uninformative training set should, on average, not perform better than 50 \% accuracy when evaluated on an uninformative test set. We create training and testing sets according to various scenarios. Each of the two classes are associated to a normal distribution, specifying the mean and variance of the sequence lengths. To generate an instance of a class, a sequence length is first drawn from its class distribution. Then each sequence element of this instance corresponds to a 300 dimensions vector (similar to GLoVe or Word2Vec word embeddings) filled with random values drawn from a $[0,1]$ uniform distribution. By selecting different means and variances, we can control the extent to which the sequence length of both classes overlaps. Overlap values used for our experiments are presented in Table~\ref{tab:exp_1_results}.

\subsection{Training Procedure} \label{subsec:training_procedure}
Unless otherwise specified, we train each classifier the same way: we generate a balanced training set of 10 000 instances (as described in Section \ref{subsec:simulated_sequence}) that we feed into a vanilla LSTM with a hidden state size of 300. Classification is made by a feedforward layer taking the last hidden state of the LSTM as input. Both LSTM and classification head are end-to-end trained together. We train the network using the Adam optimizer for ten epochs with a learning rate of 0.001. We batch the generated input in groups of 32 observations. We evaluate our models on a holdout dataset also containing 10 000 randomly generated instances. 

\subsection{Evaluation of the Problem}
This first experimentation aims to evaluate the importance of the problem by having a total control over the length distribution of the classes. We selected five sets of parameters such that the theoretical distribution of lengths overlaps by 100 \%, 80\%, 50 \%, 10\%, and 0 \%.  


We use the length distribution's overlap to quantify the source of the problem: the more similar the distributions, the less possible the input length to be used as a feature. Hence the phenomenon should be absent when the two class distributions are the same (100\% overlap) and appear gradually as the distributions move apart (0 \% overlap). We evaluate how this change in overlap affects the performance of a vanilla LSTM cell.

\subsection{Results}
The first four columns of Table \ref{tab:exp_1_results} contain the means and variances of the length distributions used for the scenarios of this experimentation. The fifth column contains the theoretical overlap \% estimated from the normal distribution p.d.f. and the scenario parameters. The last column presents the accuracy obtained with an LSTM model and random inputs. Since the sequences contain non-informative elements, all accuracy values should ideally tend to be 50 \%.

\begin{table}[h!]
    \centering
    \caption{Classifier Accuracy for Synthetic Input Scenarios}
    \begin{tabular}{cccc|c|c}
         \toprule
         $\mu_0$ & $\sigma_0$  & $\mu_1$ & $\sigma_1$  & overlap \%  &  Accuracy  \\
         \midrule
         10 & 2 & 10 & 2 & 100 \% &  0.50  \\ 
         10 & 2 & 11 & 2 & 80 \% &  0.51 \\
         10 & 2 & 13 & 3 & 50 \% &  0.71  \\
         10 & 2 & 20 & 4 & 10 \% &  0.95  \\
        10 & 2 & 100 & 10 & 0 \% &  1.00   \\
        \bottomrule
    \end{tabular}
    
    \label{tab:exp_1_results}
\end{table}

The results clearly indicate that, as the distribution overlap diminishes (i.e. as the gap between the sequence length distributions of the classes augments), the accuracy of the LSTM classifier artificially increases. We find that the problem may appear as soon as the overlapping of the two sequence length distributions is lower than 80\% and becomes overwhelming when the overlapping is 0 \%. These are statistical red flags.

\section{The Problem in Real-World Datasets}\label{sec:extent_of_the_problem}
The next experimentation aims to determine if the sequence length problem is also encountered in real-world datasets containing informative texts. For this purpose, we chose the {\tt amazon polarity} corpus \cite{zhang2015character}, a balanced dataset containing 3.6M training examples that is often used to benchmark binary classification algorithms. 

\subsection{Training Dataset Alteration}\label{sec:dataset_alter}
To conduct our experiments, we modify the original training dataset to create an artificial gap between the class length distributions. In {\tt amazon polarity}, the median lengths of negative and positive examples are 90 and 79 tokens respectively. To create a training dataset with a length gap, we remove negative examples having a length smaller than 90. Likewise, we remove positive examples with more than 79 tokens. Therefore, we reduce the length overlap to 0\% between the 2 classes, a gap that could affect the behavior of the LSTM models. With this modified training dataset, we construct LSTM models like those described in Section \ref{subsec:training_procedure}. 

\subsection{Evaluation Test Datasets}\label{sec:evaluate_the_problem_realworld}
\begin{figure}[h!]
    \centering
    \subfigure[Original Test Set]{
        \includegraphics[width=0.30\textwidth]{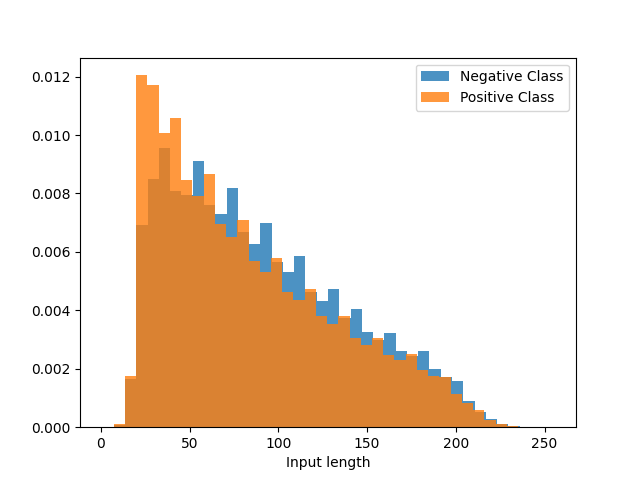}
    }
    \subfigure[Gap Test Set]{
        \includegraphics[width=0.30\textwidth]{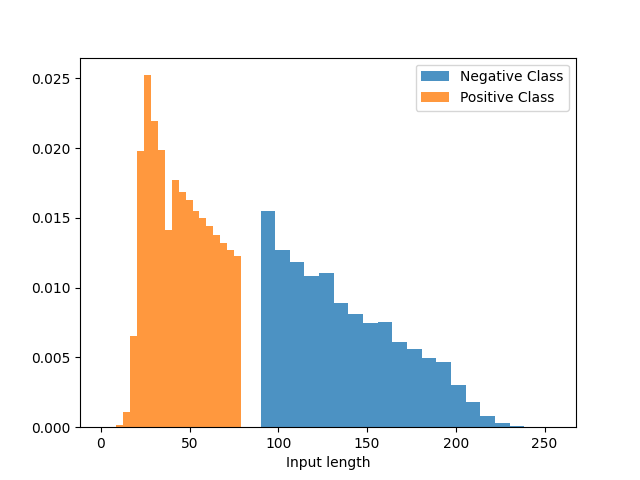}
    }
    \subfigure[Reverse Test Set]{
        \includegraphics[width=0.30\textwidth]{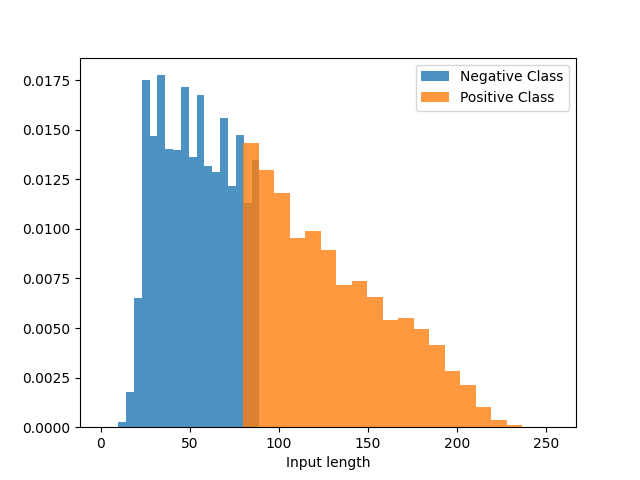}
    }
    \caption{Class length distributions}\label{fig:amazon_test_lenght_distribution}
\end{figure}

To evaluate if the model uses the unwanted length meta-feature, we alter the original test dataset with a procedure similar to what we did for the training set. However we use the different class partitions to create 2 test sets. The \textbf{Gap-test} dataset contains items with the same length distribution as the training set, i.e., $[0,79]$ for positive and $[90, \infty]$ for negative examples. The second test set, the \textbf{Reverse side}, is the complement of the \textbf{Gap-test} with length distributions of $[80,\infty]$ and $[0,89]$ for positive and negatives examples respectively. We also create a third test dataset \textbf{Reverse*} by modifying the length of the examples from the \textbf{Reverse side} dataset so they correspond to the \textbf{Gap-test}'s range. To achieve this, we truncate the long positive examples and extend the short negative examples by concatenating a duplicate of their texts. These manipulations modify the length of the examples without injecting new textual information. We present the resulting length distributions for each test dataset in Figure~\ref{fig:amazon_test_lenght_distribution}.

\subsection{Results}\label{ss:real_world_problem_evaluation}
Experimental results are presented in Table \ref{tab:amazon_problem_eval}. The first column indicates what dataset is used to train the LSTM. \textbf{Original-train} is the complete 3.6M examples of the original {\tt amazon polarity} training set, whereas the \textbf{Gap-train} corresponds to the modified training set described in Section \ref{sec:dataset_alter}. The four last columns present the accuracy score obtained on different test sets. \textbf{Original-test} refers to the original {\tt amazon polarity} test set while the other three columns refer the modified test sets presented in Section \ref{sec:evaluate_the_problem_realworld}.

\begin{table}[h!]
    \centering
    \caption{Accuracy for each Model on the Different Test Datasets}
    \begin{tabular}{c|cccc}
    \toprule
         Training Set  & Original-test & Gap-test & Reverse & Reverse* \\
          \midrule
        Original-train         & 94.7\%  & 95.1 \%  & 94.1 \% & 92.8 \% \\
        Gap-train         & \textbf{53.2} \%  & 100.0 \%  & \textbf{7.0} \%  & 90.4 \%\\
        \bottomrule
    \end{tabular}
    
    \label{tab:amazon_problem_eval}
\end{table}

We first notice that the classifier trained with the original \textbf{Original-train} dataset performs well on all the modified test datasets with results similar to the benchmarks provided in \cite{mabrouk2020deep}. However, when using the \textbf{Gap-train} for training purposes, results are significantly different from those of \textbf{Original-train} for the first three test sets. 

A important difference is observed for the \textbf{Original-test} set with a significant drop in accuracy from 94.7 \% to 53.2 \%. Even if the reviews remain unchanged, changes in class length distributions largely deteriorate the performance. It suggests that, during the training of the \textbf{Gap-train} model, some of the predictive power is moved from the textual content to the class length meta-feature. 

The \textbf{Gap-test} results indicate that the model learns extremely well when both training and test sets have the same length distribution as the accuracy reaches a perfect 100.0 \%. In contrast, for the \textbf{Reverse Side} where length distributions are reversed, performance dramatically drops to an accuracy of 7.0 \%. Two factors could contribute to this deterioration: either the \textbf{Reverse Side} examples are from a different content distribution, or the model mistakenly learned to use the input length as its prominent feature and nothing learned from other content features.

The results on the third test set \textbf{Reverse*} help us to determine the cause of this behavior. The same examples that had an accuracy of 7.0 \% now tally up to 90.4 \%. This significant improvement clearly shows us that whatever information these examples contain, the model can classify them correctly as long as training and test examples share the same length distribution. These results again support the hypothesis that recurrent cells will use input length as a prominent feature to classify documents.

A necessary conclusion is that the sequence length problem in classification tasks may result in LSTM models misperforming its actual value (100.0\% vs. 94.7\%). This performance bias poses a critical problem: comparing LSTMs with other classification algorithms could hardly be fair because the local optimum reached with sequence length at training time could provide a better performance than models relying on informative content. Finally, as our experiment shows, if the \textbf{Original} dataset had had an important gap in its class length distribution, this issue might have gone unnoticed due to the very good performance with a gap test set (as for \textbf{Gap-test}).

\section{Possible Solutions}\label{sec:results}
An intuitive approach to the sequence length problem is to address it as a regularization problem. As the network is trained to predict which class the observation belongs to, the model's ability to adjust its parameters weights is hampered by regularization and is forced to use variation-robust information. In this section, we evaluate the impact of traditional regularization techniques such as weight decay and dropout, as presented in works like  \cite{merity2017regularizing}, to mitigate our problem with the synthetic and {\tt amazon\_polarity} datasets.

\subsection{Informative Synthetic Dataset}
The regularization technique should prevent the model from learning the sequence length meta-feature while allowing salient features to be selected. In our current framework, the later part of the behavior cannot be evaluated for synthetic data as there is no content information in our completely randomly generated synthetic dataset. To overcome this limitation, we created an additional synthetic dataset containing information that could help models to discriminate between the classes.

We generate the informative training dataset with a method similar to that used for the uninformative synthetic dataset, as explained in Section \ref{subsec:simulated_sequence}. To add some informative content to a purely random dataset, we modify 10\% of the positive class sequences so they contain a filled vector of 1s. Such a feature is straightforward to learn and should lead to 100 \% classification accuracy.

\subsection{Dropout}
To evaluate if dropout can solve the sequence length problem, we add a 25 \% dropout to the synthetic inputs during training. Dropout randomly sets a percentage of inputs to zero and forces a model to use multiple features instead of a few salient ones.  

\begin{table}[h!]
    \centering
    \caption{Classification Accuracy of LSTMs Trained with Dropout using the Synthetic Datasets}
    \begin{tabular}{ccc|cc}
    \toprule
         Overlap \% & Dataset  & Dropout &  Accuracy  \\
         \midrule
         0 \% & Synthetic (random)  & 0 \% &  1.00 \\
        0 \% & Synthetic (informative) & 0 \% &  1.00 \\
         \midrule

         0 \% & Synthetic (random)  & 25 \% &  1.00 \\
         0 \% & Synthetic (informative) & 25 \% &  1.00 \\
         
         \bottomrule
    \end{tabular}
    
    \label{tab:exp_2_results_do}
\end{table}

The results presented in Table \ref{tab:exp_2_results_do} clearly indicate that the problem is not solved because the LSTM models can make perfect predictions from random content (while it should be 50\%). For both scenarios, dropout does not lower the performance of either models as compared to their performance from Table~\ref{tab:exp_1_results}. According to our experiments, using a higher dropout percentage (50 \% and 75\%) has a similar impact on both models' performance and thus should not be considered a possible solution. A possible explanation for this behavior is that dropout is applied to the observations of both classes. As the length distributions of both classes are modified proportionally by the dropout regularization, a class length overlap remains present in the inputs.

\subsection{Weight Decay on Synthetic Data}
To assess the impact of weight decay, we add a penalty proportional to the parameter values to the gradient during backpropagation. This operation forces the optimizer to converge toward a solution that uses smaller weights. 

In our problem, the random nature of our synthetic inputs requires capturing weights on all dimensions, since there is no pattern in any of the input dimensions. An optimizer regularized by weight decay prefers models with a limited number of smaller weights and should be forced to exploit the salient informative part of the vector instead of sequence lengths. 

\begin{table}[h!]
    \centering
    \caption{Classification Accuracy of LSTMs Trained with Weight Decay using the Synthetic Datasets}
    \begin{tabular}{ccc|c}
    \toprule
         Overlap \% & Dataset & Weight Decay & Accuracy \\
    \midrule
         0 \% & Synthetic (random)  & 0 & 1.00  \\
         0 \% & Synthetic (informative)    & 0 & 1.00  \\
    
    \midrule
         0 \% & Synthetic (random)  & 0.1 & 0.50  \\
         0 \% & Synthetic (informative)    & 0.1 & 1.00  \\
     \bottomrule
    \end{tabular}
    
    \label{tab:exp_2_results_wd}
\end{table}

As predicted, weight decay significantly reduces the problem of using the input sequence length as a feature. As presented in Table \ref{tab:exp_2_results_wd}, using a weight decay of 0.1 prevents LSTM models to learn the sequence length meta-feature, as the uninformative model accuracy of 50\% is equivalent to random class selection (as it should be). It is also worth noting that weight decay has no impact on the performance of the model trained with the informative synthetic dataset as the accuracy remains 100\% as it should also be. We conclude that the informative model accuracy of 100 \% (using weight decay) comes from the salient feature selection and not from the sequence length meta-feature. 

We also evaluated the impact of another implicit regularization scheme, the reduction of the LSTM hidden state size, and even with a very low amount of parameters, the LSTM learned the unwanted feature.


\subsection{Weight Decay Impact on Real-World dataset}
To verify that our conclusion from the previous section still holds for real-world data, we evaluate how weight decay modifies the performance of an LSTM model trained using the {\it amazon\_polarity} \textbf{Gap-train} dataset.

\begin{table}[h!]
    \centering
    \caption{F1 Score for Regularized and Unregularized Models Trained with the \textbf{Gap-train} Dataset}
    \begin{tabular}{c|cccc}
    \toprule
          Weight Decay & Original-test & Gap-test & Reverse & Reverse* \\
          \midrule
         0        & 53.2 \%  & 100.0 \%  & 7.0 \%  & 90.4 \%\\
         0.045 & \textbf{76.6} \%  & 89.6 \%  & \textbf{63.9} \%  & 82.3 \% \\
        \bottomrule
    \end{tabular}
    \label{tab:amazon_results}
\end{table}
As one can see, the model trained with a thoroughly searched weight decay parameter of 0.045 has a very different performance profile from its unregularized counterpart. The classification accuracy of the \textbf{Original-test} improves by 22 \%. An improvement is also observed for the \textbf{Reverse} test set, going from 7.0 \% to 63.9 \%. One should also note that the \textbf{Gap-test} performance has decreased from 100.0  \% to 89.6 \%. As we hinder the capacity of the model to learn to use the input length as features, we expect the \textbf{Gap-test} value to converge to its true value of 95.1 \% (from Table~\ref{tab:amazon_problem_eval}), obtained by training and evaluating the model on the original datasets.

These results show us that the conclusion drawn with synthetic inputs still holds for a real-world dataset, even if weight decay cannot completely negate the effect of the sequence length problem.


    


\subsection{Weight Decay Effect on Representations}
As the model learns its prediction task, parameters are fit to the data to produce document representations the classification module can separate. To demonstrate the effect of weight decay on the model, we present and analyze 2d projections (using the TSNE algorithm - see \cite{van2008visualizing}) of embeddings generated by uninformative data in models trained with and without weight decay.

\begin{figure}[hbt!]
    \centering
    
    \subfigure[Initialisation Weights]{
        \includegraphics[width=0.28\textwidth]{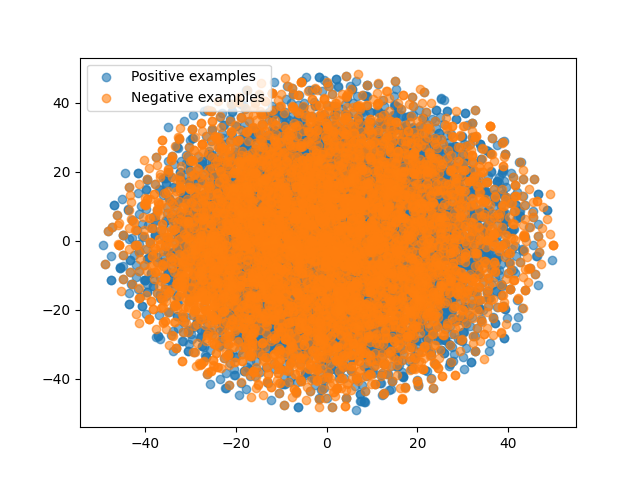} 
    }
    \subfigure[No Weight Decay]{
        \includegraphics[width=0.18\textwidth]{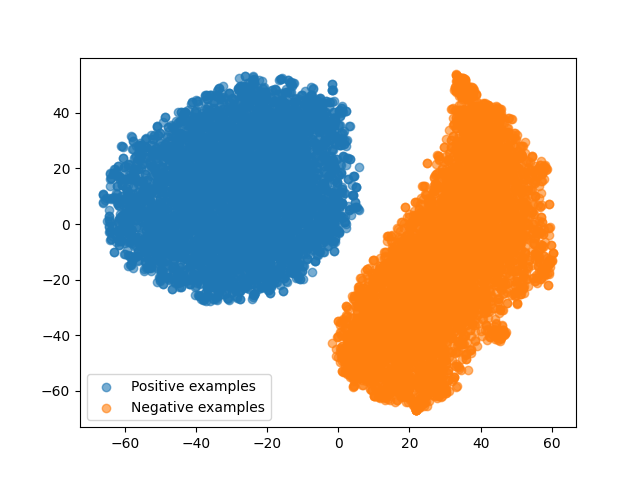}
    }\subfigure[With Weigth Decay]{
        \includegraphics[width=0.18\textwidth]{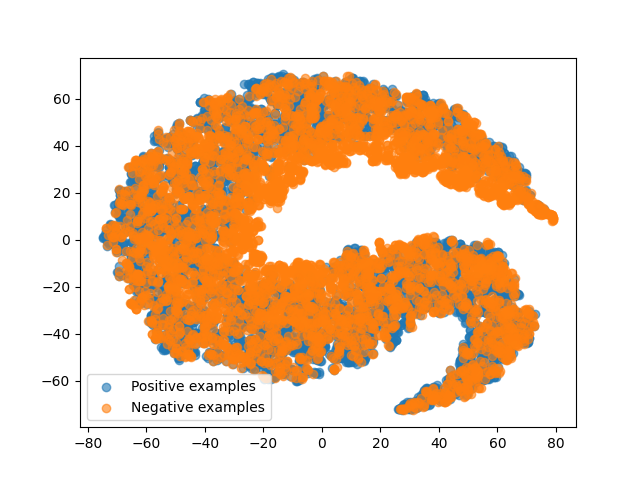}
    }
    \caption{TSNE Projections of Embeddings Generated by Models Trained with Uninformative Synthetic Data}\label{fig:tsne_uninformative}
    
\end{figure}

The scatter plots from Figure~\ref{fig:tsne_uninformative} are the embeddings (i.e. the hidden state used for classification purposes) generated by an LSTM model trained with random data. 

The first subplot \ref{fig:tsne_uninformative}(a) presents the projection of the random sequences using the initialization weights of the LSTM model. As there is not much knowledge stored in the model weights, the embeddings generated for each class completely overlap in the middle of the vector space. 

The next two scatter plots are the same random sequences passed in models trained without and with weight decay. From \ref{fig:tsne_uninformative}(b), it can be seen that the model without weight decay can easily separate the two classes even though no informational content can support this result except the sequence length meta-feature. However, as shown in \ref{fig:tsne_uninformative}(c), applying weight decay removes this class separation and causes the generated embeddings to overlap once again.

This figure support our hypothesis that LSTM models can learn meta features such as sequence lengths and that weight decay helps to alleviate this problem.

\section{Conclusion}
In conclusion, this paper describes a problem encountered in our current research work: recurrent networks tend to use the difference in class length distributions as a prominent classification feature instead of learning the important textual features. This problem is not trivial because it makes recurrent models fragile to concept drift and can lead to useless explanations as for their decisions. Moreover, although using the sequence length difference helps the model converge to a local optimum, it provides misleading performance that could lead to an unfair comparison between classification algorithms of different types.

To isolate the problem, we evaluated its significance using engineered synthetic examples. Using LSTM models, sequence length issues appear as soon as the overlap between the class length distribution is lower than 80\% and would trivialize the classification task whenever it approaches 0 \%. We also demonstrated that the undesired length meta-feature can also become predominant in a real-world dataset affected by class length imbalance.

We then showed and explained that weight decay regularization can mitigate the impact of the problem on both synthetic and real-world training datasets. Even if weight decay could fully not hinder the effect of the sequence length problem, it opens research opportunities in artificial example augmentation, where training sets could be enhanced to include examples that would balance the input length distribution. 

Preliminary results assert that more advanced classification models, such as hierarchical attention networks \cite{yang2016hierarchical} and transformers \cite{vaswani2017attention}, are also impacted by this problem. Future work is required to evaluate the extent of the problem on these architectures, assess its impact on classification tasks and determine whether weight decay regularization also solves the problem as much as it does for LSTM architectures.

\bibliographystyle{apalike}
\bibliography{references}

\begin{thebibliography}{}

\bibitem[Beltagy et~al., 2020]{beltagy2020longformer}
Beltagy, I., Peters, M.~E., and Cohan, A. (2020).
\newblock Longformer: The long-document transformer.
\newblock {\em arXiv preprint arXiv:2004.05150}.

\bibitem[Fang et~al., 2018]{fang2018can}
Fang, M., Zhou, Z., Chen, S., and McClelland, J. (2018).
\newblock Can a recurrent neural network learn to count things?
\newblock In {\em CogSci}, pages 360--365.

\bibitem[Gers and Schmidhuber, 2000]{gers2000recurrent}
Gers, F.~A. and Schmidhuber, J. (2000).
\newblock Recurrent nets that time and count.
\newblock In {\em Proceedings of the IEEE-INNS-ENNS International Joint
  Conference on Neural Networks. IJCNN 2000. Neural Computing: New Challenges
  and Perspectives for the New Millennium}, volume~3, pages 189--194. IEEE.

\bibitem[Hochreiter and Schmidhuber, 1997]{hochreiter1997long}
Hochreiter, S. and Schmidhuber, J. (1997).
\newblock Long short-term memory.
\newblock {\em Neural computation}, 9(8):1735--1780.

\bibitem[Mabrouk et~al., 2020]{mabrouk2020deep}
Mabrouk, A., Redondo, R. P.~D., and Kayed, M. (2020).
\newblock Deep learning-based sentiment classification: a comparative survey.
\newblock {\em IEEE Access}, 8:85616--85638.

\bibitem[Merity et~al., 2017]{merity2017regularizing}
Merity, S., Keskar, N.~S., and Socher, R. (2017).
\newblock Regularizing and optimizing lstm language models.
\newblock {\em arXiv preprint arXiv:1708.02182}.

\bibitem[Pascanu et~al., 2013]{pascanu2013difficulty}
Pascanu, R., Mikolov, T., and Bengio, Y. (2013).
\newblock On the difficulty of training recurrent neural networks.
\newblock In {\em International conference on machine learning}, pages
  1310--1318. PMLR.

\bibitem[Rodriguez et~al., 1999]{rodriguez1999recurrent}
Rodriguez, P., Wiles, J., and Elman, J.~L. (1999).
\newblock A recurrent neural network that learns to count.
\newblock {\em Connection Science}, 11(1):5--40.

\bibitem[Van~der Maaten and Hinton, 2008]{van2008visualizing}
Van~der Maaten, L. and Hinton, G. (2008).
\newblock Visualizing data using t-sne.
\newblock {\em Journal of machine learning research}, 9(11).

\bibitem[Vari{\v{s}} and Bojar, 2021]{varivs2021sequence}
Vari{\v{s}}, D. and Bojar, O. (2021).
\newblock Sequence length is a domain: Length-based overfitting in transformer
  models.
\newblock {\em arXiv preprint arXiv:2109.07276}.

\bibitem[Vaswani et~al., 2017]{vaswani2017attention}
Vaswani, A., Shazeer, N., Parmar, N., Uszkoreit, J., Jones, L., Gomez, A.~N.,
  Kaiser, {\L}., and Polosukhin, I. (2017).
\newblock Attention is all you need.
\newblock {\em Advances in neural information processing systems}, 30.

\bibitem[Yang et~al., 2016]{yang2016hierarchical}
Yang, Z., Yang, D., Dyer, C., He, X., Smola, A., and Hovy, E. (2016).
\newblock Hierarchical attention networks for document classification.
\newblock In {\em Proceedings of the 2016 conference of the North American
  chapter of the association for computational linguistics: human language
  technologies}, pages 1480--1489.

\bibitem[Zhang et~al., 2015]{zhang2015character}
Zhang, X., Zhao, J., and LeCun, Y. (2015).
\newblock Character-level convolutional networks for text classification.
\newblock {\em Advances in neural information processing systems}, 28.

\end{thebibliography}
\end{document}